\documentclass[10pt]{article}
\usepackage[english]{babel}
\usepackage[a4paper,textwidth=122mm,textheight=193mm]{geometry}
\usepackage{url}
\usepackage{graphicx}
\usepackage{multirow}
\usepackage{calc}
\usepackage{booktabs}

\begin{document}
\title{Interpretable Aircraft Engine Diagnostic via Expert Indicator Aggregation\footnote{This study is supported by a grant from Snecma, Safran Group, one of the world’s leading manufacturers of aircraft and rocket engines, see
 \protect\url{http://www.snecma.com/} for details.}}
\author{Tsirizo Rabenoro${}^{\dagger}$, J\'{e}r\^{o}me Lacaille${}^{\dagger}$,
  Marie Cottrell${}^{\star}$\\and Fabrice Rossi${}^{\star,\bullet}$\\[1em]
\small $\dagger$ Snecma, Groupe Safran, 77550 Moissy Cramayel, France\\
\small $\star$ SAMM (EA 4543), Universit\'{e} Paris 1,\\
\small 90, rue de Tolbiac, 75634 Paris Cedex 13, France\\
\small $\bullet$ corresponding author, \url{Fabrice.Rossi@univ-paris1.fr}}
\date{}
\maketitle
\begin{abstract}
  Detecting early signs of failures (anomalies) in complex systems is one of
  the main goal of preventive maintenance. It allows in particular to avoid
  actual failures by (re)scheduling maintenance operations in a way that
  optimizes maintenance costs. Aircraft engine health monitoring is one
  representative example of a field in which anomaly detection is
  crucial. Manufacturers collect large amount of engine related data during
  flights which are used, among other applications, to detect anomalies.  This
  article introduces and studies a generic methodology that allows one to
  build automatic early signs of anomaly detection in a way that builds upon
  human expertise and that remains understandable by human operators who make
  the final maintenance decision. The main idea of the method is to generate a
  very large number of binary indicators based on parametric anomaly scores
  designed by experts, complemented by simple aggregations of those scores. A
  feature selection method is used to keep only the most discriminant
  indicators which are used as inputs of a Naive Bayes classifier. This give
  an interpretable classifier based on interpretable anomaly detectors whose
  parameters have been optimized indirectly by the selection process. The
  proposed methodology is evaluated on simulated data designed to reproduce
  some of the anomaly types observed in real world engines.
\end{abstract}

\section{Introduction}
Automatic anomaly detection is a major issue in numerous areas
and has generated a vast scientific literature
\cite{chandola2009anomaly}. We focus in this paper on a very important
application case, aircraft engine health monitoring which aims at detecting
early signs of failures, among other applications
\cite{asheraryan2013surveyhealth,tumer1999survey}. 
Aircraft engines are generally made extremely reliable by their conception
process and thus have low rate of operational events.  For example, in 2013,
the CFM56-7B engine, produced jointly by Snecma and GE aviation, has a
rate of in flight shut down (IFSD) of $0.002$ (per 1000 Engine Flight Hour) and
a rate of aborted take-off (ATO) of $0.005 $ (per 1000 departures).  This
dispatch availability of nearly 100 \% (99.962 \% in 2013) is obtained
via regular maintenance operations but also via engine health monitoring (see
also e.g. \cite{vasov2007reliability} for an external evaluation). 

This monitoring is based, among other sources, on data transmitted by
satellites\footnote{using the commercial standard Aircraft Communications
  Addressing and Reporting System (ACARS, see
  \url{http://en.wikipedia.org/wiki/ACARS}), for instance.}  between aircraft
and ground stations. Typical transmitted messages include engine status
overview as well as useful measurements collected as specific instants (e.g.,
during engine start). Flight after flight, measurements sent are analyzed in
order to detect anomalies that are early signs of degradation.  Potential
anomalies can be automatically detected by algorithms designed by experts.  If
an anomaly is confirmed by a human operator, a maintenance recommendation is
sent to the company operating the engine.

As a consequence, unscheduled inspections of the engine are sometimes
required. These inspections are due to the abnormal measurements.  Missing a
detection of early signs of degradation can result in an IFSD, an ATO or a
delay and cancellation (D\&C).  Despite the rarity of such events, companies
need to avoid them to minimize unexpected expenses and customers'
disturbance. Even in cases where an unscheduled inspection does not prevent
the availability of the aircraft, it has an attached cost: it is therefore
important to avoid as much as possible useless inspections.

We describe in this paper a general methodology to built complex automated
decision support algorithms in a way that is comprehensible by human
operators who take final decisions. The main idea of our approach is to leverage expert knowledge in
order to build hundreds of simple binary indicators that are all signs of the
possible existence of an early sign of anomaly in engine health monitoring data. The
most discriminative indicators are selected by a standard forward feature
selection algorithm. Then an automatic classifier is built on those
features. While the classifier decision is taken using a complex decision
rule, the interpretability of the features, their expert based nature and
their limited number allows the human operator to at least partially
understand how the decision is made. It is a requirement to have a trustworthy
decision for the operator. It should be noted that while this paper focuses on
aircraft engines, the methodology can be applied to various other
contexts. For instance a related but simpler methodology was proposed
in \cite{hegedus2011methodology} in the context of malware detection. 

The rest of the paper is organized as follows. Section \ref{sec:context}
describes the engine health monitoring context. The methodology is presented in
details in Section \ref{sec:arch-decis-proc}.  Section
\ref{sec:simulation-study} is dedicated to a simulation study that validates
the proposed approach.

\section{Application context}\label{sec:context}
\subsection{Flight data}
\begin{figure}
  \begin{center}
    \includegraphics[width=0.6\linewidth]{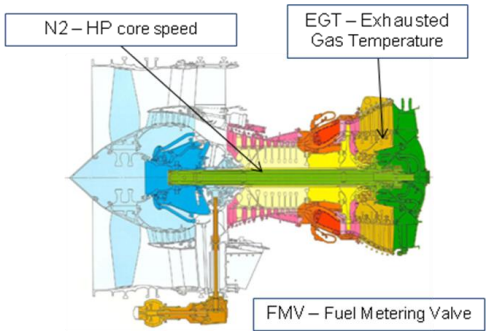}
  \end{center}
  \caption{Localization of some followed parameters on the Engine}
  \label{fig:startsequence}
\end{figure}

Engine health monitoring is based in part on flight data acquisition. Engines
are equipped with multiple sensors which measure different physical quantities
such as the high pressure core speed (N2), the Fuel Metering Valve (FMV), the
Exhausted Gas Temperature (EGT), etc. (See Figure \ref{fig:startsequence}.)
Those measures are monitored in real time during the flight. For instance the
quantities mentioned before (N2, FMV, etc.) are analyzed, among others, during
the engine starting sequence. This allows one to check the good health of the
engine.  If a potential anomaly is detected, a diagnostic is sent to an
operator of the owner of the engine. The airline may then have to postpone the
flight or cancel it, depending on the criticality of the fault and the
estimated repair time.

The monitoring can also be done flight after flight to detect any change
that can be flagged as early signs of degraded behavior. This is done by
compressing the in flight measurements into engine status overviews. The methodology
introduced in this article is mostly designed for this latter kind of monitoring.

\subsection{Detecting faults and abnormal behaviors}
Traditional engine health monitoring is strongly based on expert knowledge and
field experience (see e.g. \cite{asheraryan2013surveyhealth,tumer1999survey}
for surveys and \cite{flandrois2009expertise} for a concrete example).
Faults and early signs of faults are identified from suitable measurements
associated to adapted computational transformation of the data. For instance,
the different measurements (temperatures, vibration, etc.) are influenced by
the flight parameters (e.g. throttle position) and conditions (outside
temperature, etc.). Variations in the measured values can therefore result
from variations in the parameters and conditions rather than being due to
abnormal behavior. Thus a typical computational transformation consists in
preprocessing the measurements in order to remove dependency to the flight
context \cite{lacaille2009standardized}.

In practice, the choice of measurements and computational transformations is
generally done based on expert knowledge.  For instance in
\cite{rabenoroinstants}, a software is designed to record expert decision
about a time interval on which to monitor the evolution of such a measurement
(or a time instant when such a measurement should be recorded). Based on the
recorded examples, the software calibrates a pattern recognition model that
can automatically reproduce the time segmentation done by the expert. Once the
indicators have been computed, the normal behavior of the indicators can be
learned. The residuals between predictions and actual indicators can be
statistically modeled as a Gaussian vector, for instance. A score measurement is
obtained from the likelihood of this distribution. The normalized vector is a
failure score signature that may be described easily by experts to identify
the fault origin, in particular because the original indicators have some
meaning for them. See \cite{come2010aircraft}, \cite{flandrois2009expertise}
and \cite{lacaille2009maturation} for other examples.

However experts are generally specialized on particular subsystems, thus each
algorithm focuses mainly on a specific subsystem despite the need of a
diagnostic of the whole system.

\subsection{Data and detection fusion}
The global diagnostic is currently done by the operator who collects all
available results of diagnostic applications. The task of taking a decision
based on all incoming information originating from different subsystems is
difficult. A first difficulty comes from dependencies between subsystems which
means that for instance in some situations, a global early sign of failure could be
detected by discovering discrepancies between seemingly perfectly normal
subsystems. In addition, subsystem algorithms can provide conflicting results
or decisions with very low confidence levels.  Furthermore, the extreme
reliabilities of engines lead to an exacerbated trade off between false alarm
levels and detection levels, leading in general to a rather high level of
false alarms, at least at the operator level. Finally, the role of the
operator is not only to identify a possible early sign of failure, but also to
issue recommendations on the type of preventive maintenance needed. In other
words, the operator needs to identify the possible cause of the potential
failure. 

\subsection{Objectives}
The long term goal of engine health monitoring is to reach automated accurate,
trustworthy and precise maintenance decisions during optimally scheduled shop
visits, but also to drastically reduce operational events such as IFSD and
ATO.  However, partly because of the current industrial standard, pure black
box modeling is unacceptable. Indeed, operators are currently trained to
understand expertly designed indicators and to take complex integrated
decisions on their own. In order for a new methodology to be accepted by
operators, it has at least to be of a gray box nature, that is to be
(partially) explainable via logical and/or probabilistic reasoning. Then, our
objective is to design a monitoring methodology that helps the human operator
by proposing integrated decisions based on expertly designed indicators with a
``proof of decision''.

\section{Architecture of the Decision Process}\label{sec:arch-decis-proc}
\subsection{Engine health monitoring data}\label{sec:engine-health-monit}
In order to present the proposed methodology, we first describe the data
obtained via engine health monitoring and the associated decision problem. 

We focus here on ground based long term engine health monitoring. Each flight
produces dozens of timestamped flight events and data. Concatenating those
data produces a multivariate temporal description of an engine whose
dimensions are heterogeneous. In addition, sampling rates of individual
dimensions might be different, depending on the sensors, the number of
critical time points recorded in a flight for said sensor, etc.

Based on expert knowledge, this complex set of time series is turned into a
very high dimensional indicator vector. The main idea, outlined in the
previous section, is that experts generally know what is the expected behavior
of a subsystem of the engine during each phase of the flight. Then the
dissimilarity between the expected behavior and the observed one can be
quantified leading to one (or several) anomaly scores. Such scores are in turn
transformed into binary indicators where $1$ means an anomaly is detected and
$0$ means no anomaly detected. This is somewhat related to the way malware are
characterized in \cite{hegedus2011methodology}, but with a more direct
interpretation of each binary feature\footnote{In
  \cite{hegedus2011methodology}, a value $1$ for a feature simply means that
  the software under study has a particular quality associated to the feature,
  without knowing whether this quality is an indication of its malignity.}.

This transformation has two major advantages: it homogenizes the data and
it introduces simple but informative features (each indicator is associated to a
precise interpretation related to expert knowledge). It leads also to a loss
of information as the raw data are in general not recoverable from the
indicators. This is considered here a minor inconvenience as long as the
indicators capture all possible failure modes. This will be partially
guaranteed by including numerous variants of each indicator (as explained
below). On a longer term, our approach has to be coupled with field experience
feedback and expert validation of its coverage.

After the expert guided transformation, the monitoring problem becomes a rather
standard classification problem: based on the binary indicators, the decision
algorithm has to decide whether there is an anomaly in the engine and if, this
is the case, to identify the type of the anomaly (for instance by identifying
the subsystem responsible for the potential problem).

We describe now in more details the construction of the binary indicators. 

\subsection{Some types of anomalies}\label{sec:some-types-anomalies}
\begin{figure}
\begin{center}
\scriptsize
\begin{minipage}[c]{0.32\textwidth}
    \includegraphics[width=\linewidth]{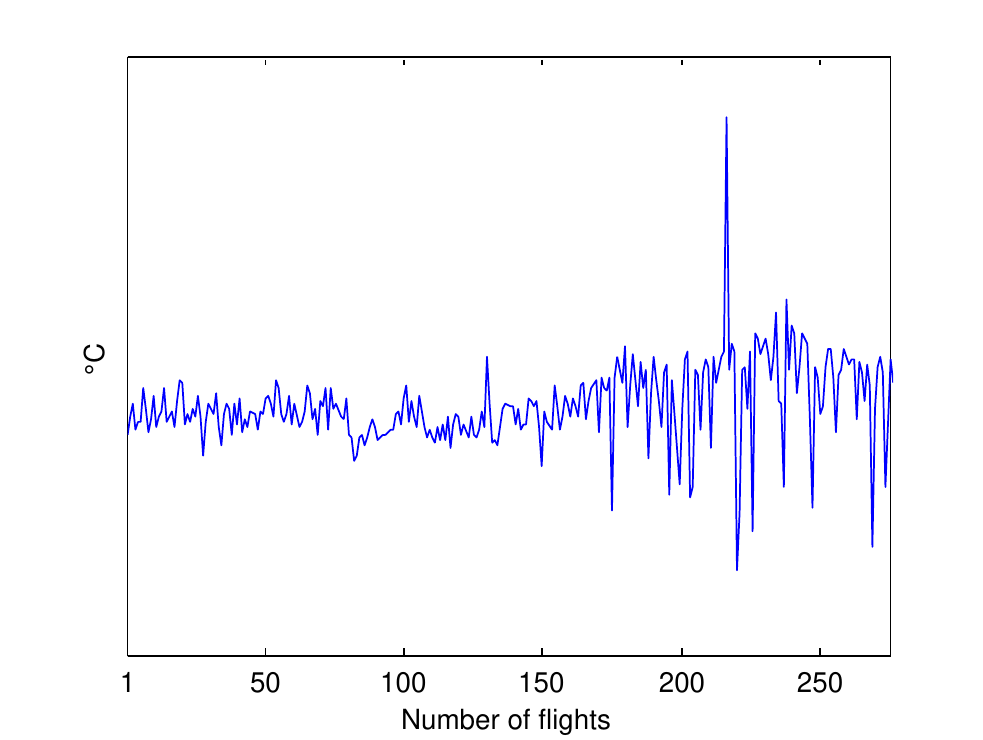}
    \caption{\scriptsize Variance shift}
   \label{fig:var1}
\end{minipage}
\begin{minipage}[c]{0.32\textwidth}
    \includegraphics[width=\linewidth]{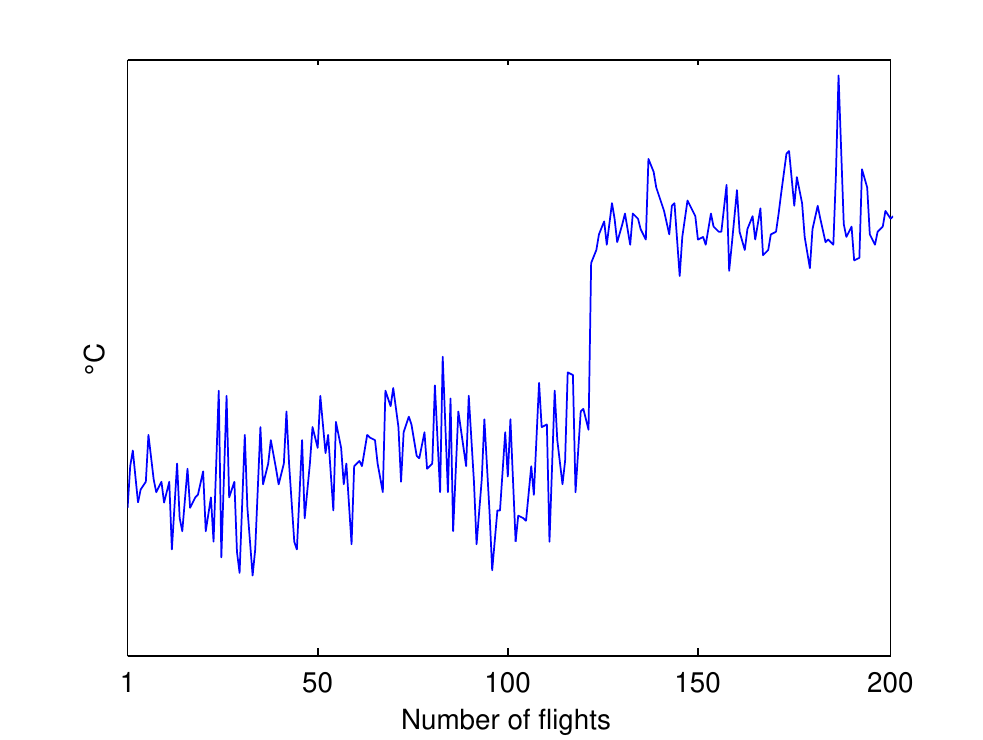}
    \caption{\scriptsize Mean shift}
   \label{jump1}
\end{minipage}
\begin{minipage}[c]{0.32\textwidth}
    \includegraphics[width=\linewidth]{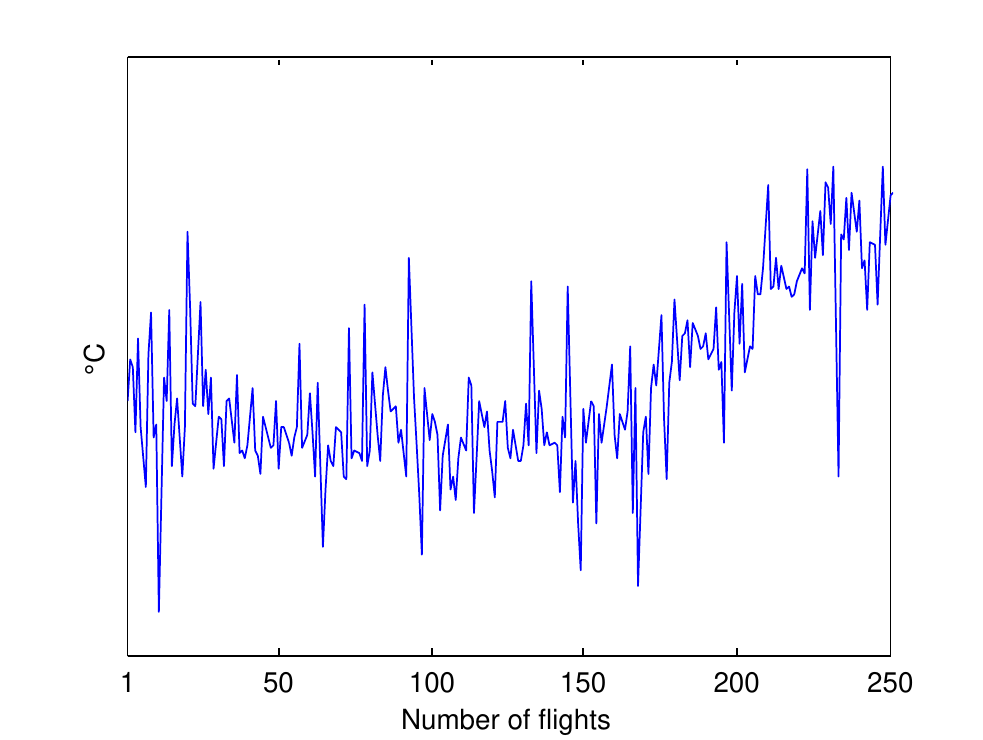}
    \caption{\scriptsize Trend modification}
   \label{trend1}
\end{minipage}
\normalsize
\end{center}
\end{figure}

Some typical univariate early signs of anomalies are shown on Figures \ref{fig:var1},
\ref{jump1} and \ref{trend1} which display the evolution through time of a
numerical value extracted from real world data. One can identify,
with some practice, a variance shift on Figure \ref{fig:var1}, a mean shift
on Figure \ref{jump1} and a trend modification (change of slope) on Figure
\ref{trend1}. In the three cases, the change instant is roughly at the center
of the time window. 

The main assumption used by experts in typical situations is that, when
external sources of change have been accounted for, the residual signal should
be stationary in a statistical sense. Assuming that a signal of length $m$ 
\[
X_{i} =(X_{i1}(\theta_1), ..., X_{im}(\theta_m)),
\]
is made of $m$ values generated independently from a fixed parametric
probability model with a parameter $\theta$. In this framework, the signal is
stationary if the parameters are fixed, that is, if all the $\theta_j$ are
identical. Then, detecting an anomaly amounts to detecting a change in the
time series (as illustrated by the three Figures above). This can be done via
numerous techniques \cite{chandola2009anomaly} in particular with well known
statistical tests \cite{basseville1993detection}. In the multivariate cases,
similar shifts in the signal can be associated to anomalies. More complex
scenarios, involving for instance time delays, can also be designed by
experts.

\subsection{Exploring the parameter space}\label{sec:from-anomaly-types}
While experts can generally describe explicitly what type of change they are
expecting for some specific early signs of anomaly, they can seldom provide
detailed parameter settings for statistical tests (or even for the aggregation
technique that could lead to a statistical test after complex
calculations). To maximize coverage it seems natural to include numerous
indicators based on variations of the anomaly detectors compatible with expert
knowledge.

Let us consider for illustration purpose that the expert recommends to look
for shifts in mean of a certain quantity as early signs of a specific
anomaly. If the expert believes the quantity to be normally distributed with a
fixed variance, then a natural test would be Student's t-test. If the expert
has no strong priors on the distribution, a natural test would be the
Mann–Whitney U test. Both can be included to maximize coverage.

Then, in both cases, one has to assess the time scale of the shift. Indeed those
tests work by comparing summary statistics of two populations, before and
after a possible change point. To define the populations, the expert has to
specify the length of the time windows to consider before and after the
possible change point: this is the expected time scale at which the shift will
appear. In most cases, the experts can only give a rough idea of the
scale. Again, maximizing the coverage leads to the inclusion of several scales
compatible with the experts' recommendations. 

Given the choice of the test, of its scale and of a change point, one can
construct a statistic. A possible choice for the indicator could be the 
value of the statistic or the associated $p$-value. However, we choose to use simpler indicators
to ease their interpretation. Indeed, the raw value of a statistic is
generally difficult to interpret. A $p$-value is easier to understand because
of the uniform scale, but can still lead to misinterpretation by operators
with insufficient statistical training. We therefore choose to use binary
indicators for which the value 1 corresponds to a rejection of the null
hypothesis of the underlying test to a given level (the null hypothesis is
here the case with no mean shift). 

\subsection{Confirmatory indicators}\label{sec:conf-indic}
Finally, as pointed out before, aircraft engines are extremely reliable, a fact
that increases the difficulty in balancing sensibility and specificity of anomaly
detectors. In order to alleviate this difficulty, we build high level
indicators from low level tests. For instance, if we monitor the evolution of
a quantity on a long period compared to the expected time scale of anomalies,
we can compare the number of times the null hypothesis of a test has been
rejected on the long period with the number of times it was not rejected, and
turn this into a binary indicator with a majority rule. 

\subsection{Decision}\label{sec:decision}
To summarize, we construct parametric anomaly scores from expert knowledge,
together with acceptable parameter ranges. By exploring those ranges, we
generate numerous (possible hundreds of) binary indicators. Each indicator can be
linked to an expertly designed score with a specific set of parameters and
thus is supposedly easy to interpret by operators. Notice that while we are
focused in this presentation on temporal data, this framework can be applied
to any data source. 

The final decision step consists in classifying these high dimensional binary
vectors into at least two classes, i.e., the presence or absence of an
anomaly. A classification into more classes is highly desirable if possible,
for instance to further discriminate between seriousness of anomalies and/or
sources (in terms of subsystems of the engine). 

As explained before, we aim in the long term at gray box modeling, so while
numerous classification algorithms are available see
e.g. \cite{kotsiantis2007supervised}, we shall focus on interpretable ones. In
this paper, we choose to use Random Forests \cite{breiman2001random} as they
are very adapted to binary indicators and to high dimensional data. They are
also known to be robust and to provide state-of-the-art classification
performances at a very small computational cost. While they are not as
interpretable as their ancestors CART \cite{breiman1984classification}, they
provide at least variable importance measures that can be used to identify the
most important indicators. 

Another classification algorithms used in this paper is naive Bayes classifier
\cite{koller2009probabilistic} which is also appropriate for high dimensional
data.  They are known to provide good results despite the strong assumption of
the independence of features given the class. In addition, decisions taken by
a naive Bayes classifier are very easy to understand thanks to the estimation
of the conditional probabilities of the feature in each class. Those
quantities can be shown to the human operator as references. 

Finally, while including hundreds of indicators is important to give a broad
coverage of the parameter spaces of the expert scores and thus to maximize the
probability of detecting anomalies, it seems obvious that some redundancy will
appear. Unlike \cite{hegedus2011methodology} who choose features by random
projection, the proposed methodology favors interpretable solutions, even at
the expense of the classification accuracy: the goal is to help the human
operator, not to replace her/him. Therefore, we have chosen to apply a feature
selection technique \cite{guyon2003introduction} to this problem. The
reduction of number of features will ease the interpretation by limiting the
quantity of information transmitted to the operators in case of a detection by
the classifier. Among the possible solutions, we choose to use the Mutual
information based technique Minimum Redundancy Maximum Relevance (mRMR,
\cite{peng2005feature}) which was reported to give excellent results on high
dimensional data (see also \cite{fleuret-2004} for another possible choice).

\section{A simulation study}\label{sec:simulation-study}

\subsection{Introduction}
It is difficult to find real data with early signs of degradations, 
because their are scarce and moreover the scheduled maintenance operations tend to remove
these early signs.  Experts could study in detail recorded data to find early
signs of anomalies whose origins were fixed during maintenance but it is close
to looking for a needle in a haystack, especially considering the huge amount
of data to analyze. We will therefore rely in this paper on
simulated data. Our goal is to validate the interest of the proposed
methodology in order to justify investing in the production of carefully
labelled real world data. 

In this section we begin by the description of the simulated data used for the
evaluation of the methodology, and then we will present the performance
obtained on this data.

\subsection{Simulated data}
The simulated data are generated according to the univariate shift models
described in Section \ref{sec:some-types-anomalies}: each observation $X_i$ is a
short time series which is recorded as at specific time points, e.g.,
$X_i=(X_i(t_{ij}))_{1\leq j\leq m_i}$. As pointed out in Section
\ref{sec:engine-health-monit}, signals can have different time
resolutions. This difficulty is integrated in the simulated data by using
different lengths/dimensions for each observation (hence the $m_i$ numbers of
time points). Notice that the time points $(t_{ij})_{1\leq j\leq m_i}$ are
introduced here only for generative purposes and are not used in the actual
decision process. For multivariate data sets, they could become useful
(e.g. to correlate potential shift detection), but this is out of the scope of
the present paper. 

In the rest of the paper, the notation $Z\sim \mathcal{N}(\mu,\sigma^2)$ says
that the random variable $Z$ follows a Gaussian distribution with mean $\mu$
and variance $\sigma^2$ and the notation $W\sim \mathcal{U}(S)$ says that
the random variable $W$ follows the uniform distribution on the set $S$ (which
can be finite such as $\{1, 2, 3\}$ or infinite such as $[0, 1]$).

\begin{figure}[htbp]
\begin{center}
\scriptsize
\begin{minipage}[c]{0.32\textwidth}
    \includegraphics[width=\linewidth,height=0.575\linewidth]{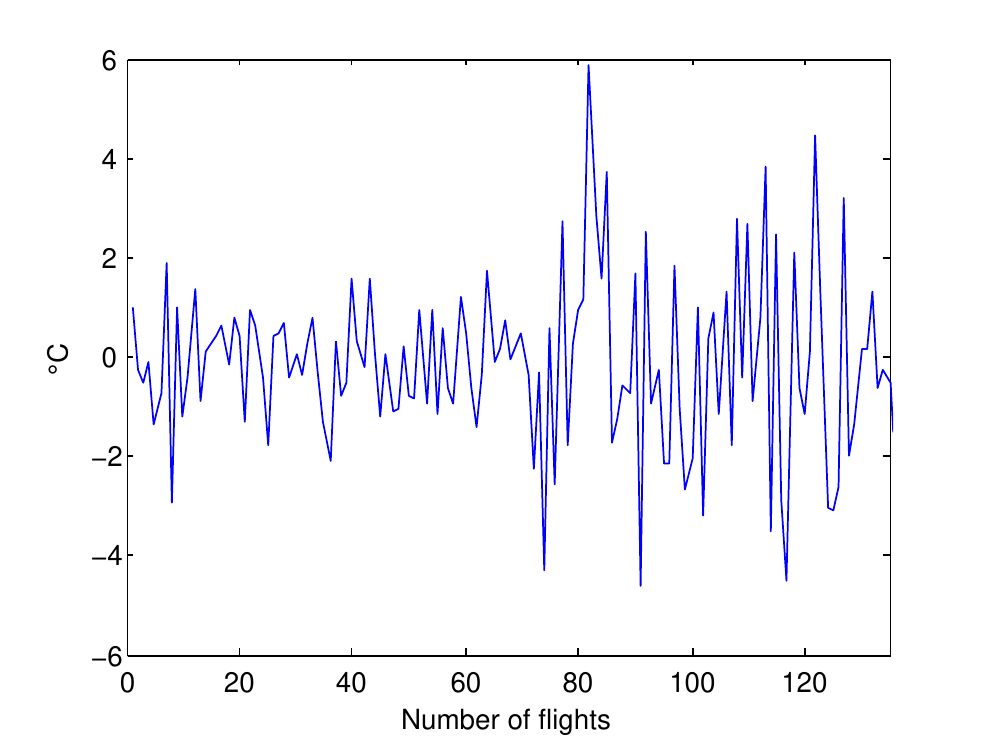}
    \caption{\scriptsize variance shift}
   \label{Fig:var2}
\end{minipage}
\begin{minipage}[c]{0.32\textwidth}
    \includegraphics[width=\linewidth,height=0.575\linewidth]{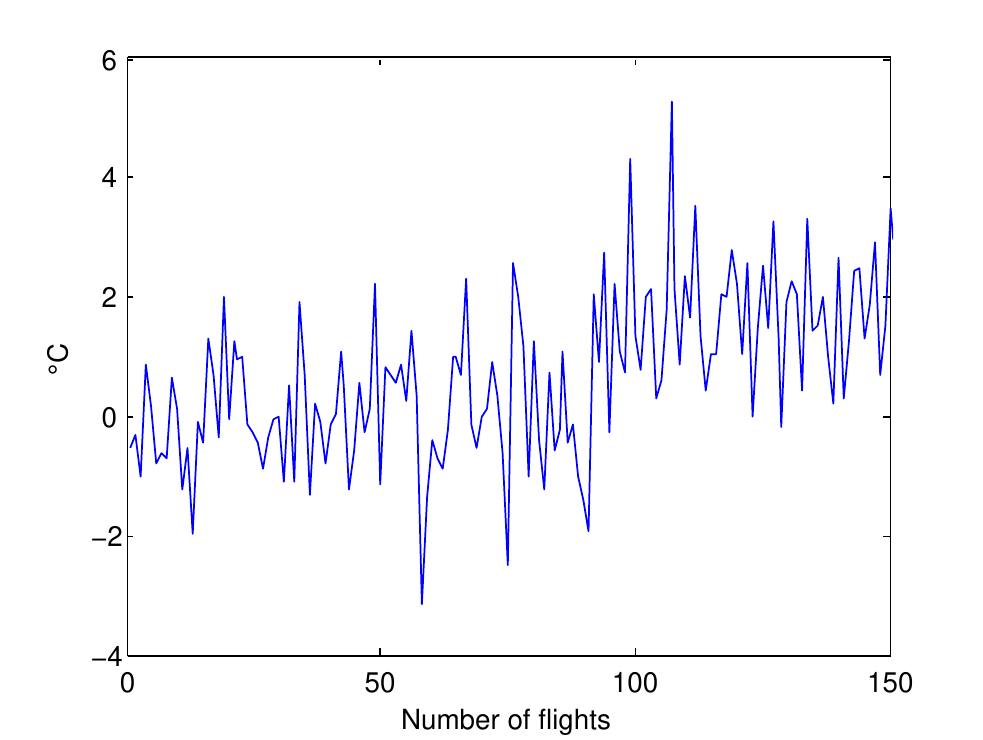}
    \caption{\scriptsize mean shift}
   \label{Fig:jump2}
\end{minipage}
\begin{minipage}[c]{0.32\textwidth}
    \includegraphics[width=\linewidth,height=0.575\linewidth]{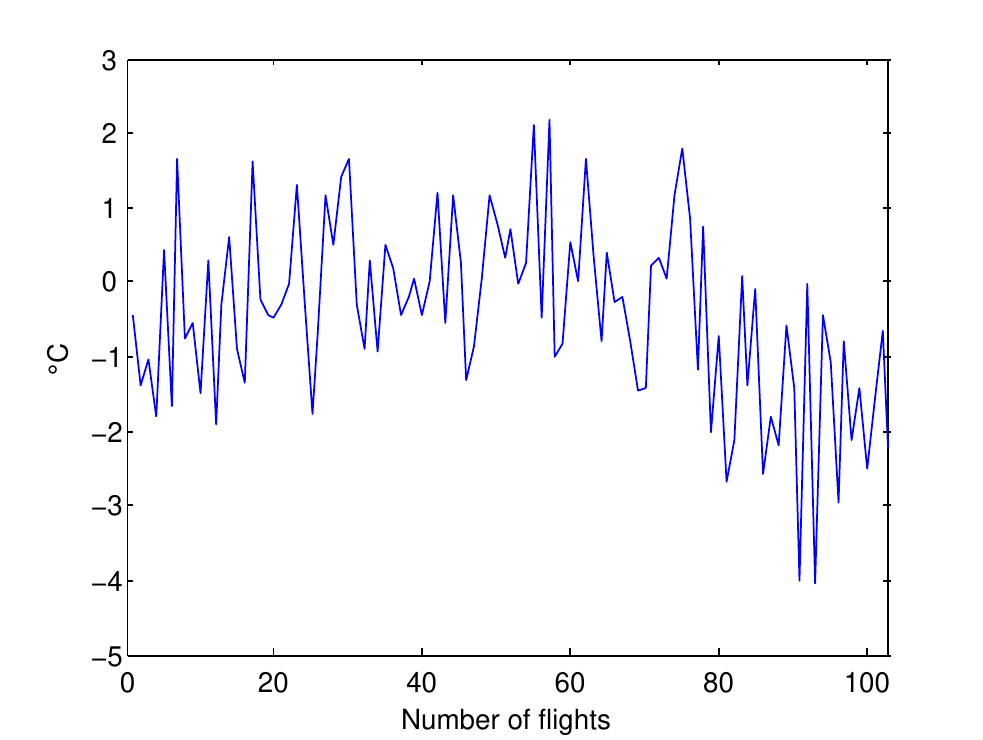}
    \caption{\scriptsize trend modification}
   \label{Fig:trend2}
\end{minipage}
 \normalsize
\end{center}
\end{figure}

We generate two data sets: a simple one $A$ and a slightly more complex one
$B$. In both cases, it is assumed that expert based normalisation has been
performed. Therefore when no shift in the data distribution occurs, we observe
a stationary random noise modeled by the standard Gaussian distribution. 
This assumption made about the noise may seem very strong but the actual goal of this paper is to evaluate the methodology and we choose to use simple distribution and simple statistical tests.
In the future, we plan to use more realistic noise associated with more complex tests.
 An observation $X_i$ with no shift is then generated as follows:
\begin{enumerate}
\item $m_i$, the length of the signal, is chosen as
  $m_i\sim\mathcal{U}(\{100,101, \ldots,200\})$ ;
\item the $m_i$ values $(X_i(t_{ij}))_{1\leq j\leq m_i}$ are sampled
  independently from the standard Gaussian distribution, that is
  $X_i(t_{ij})\sim \mathcal{N}(\mu=0,\sigma^2=1)$.
\end{enumerate}
Anomalous signals use the same distribution of the signal length as normal
signals. More precisely, an anomalous observation $X_i$ is generated as
follows:
\begin{enumerate}
\item $m_i$, the length of the signal, is chosen as
  $m_i\sim\mathcal{U}(\{100,101, \ldots,200\})$ ;
\item the change point $t^s_i$ is chosen as 
\[
t_i^s\sim\mathcal{U}\left(\left\{\left\lfloor\frac{2m_i}{10}\right\rfloor,\left\lfloor\frac{2m_i}{10}\right\rfloor+1,\ldots,\left\lfloor\frac{8m_i}{10}\right\rfloor\right\}\right),
\]
where $\lfloor x\rfloor$ is the integer part of $x$. For instance, if $m_i=100$, then the change
point is chosen among the time points $t_{20}$ to $t_{80}$ ;
\item the $m_i$ values $(X_i(t_{ij}))_{1\leq j\leq m_i}$ are generated
  according to one anomaly model. 
\end{enumerate}
Anomalies are in turn modelled after the three examples given in Figures
\ref{fig:var1}, \ref{jump1} and \ref{trend1}. The three types of shift are:
\begin{enumerate}
\item a variance shift: in this case, the parameter of the shift is the
  variance after the change point, $\sigma^2_i$, with
  $\sigma_i\sim\mathcal{U}([1.01, 5])$. Given $\sigma_i$, the
  $(X_i(t_{ij}))_{1\leq j\leq 
    m_i}$ are sampled independently as $X_i(t_{ij})\sim\mathcal{N}(\mu=0,\sigma^2=1)$ when
  $t_{ij}< t^s_i$ (before the change point) and 
  $X_i(t_{ij})\sim\mathcal{N}(\mu=0,\sigma^2=\sigma_i^2)$ when $t_{ij}\geq t^s_i$ (after the
  change point). See Figure \ref{Fig:var2} for an example;
\item a mean shift: in this case, the parameter of the shift is the mean after
  the change point, $\mu_i$. In set $A$, $\mu_i\sim\mathcal{U}([1.01, 5])$
  while in set $B$, $\mu_i\sim\mathcal{U}([0.505, 2.5])$. Given $\mu_i$, the
  $(X_i(t_{ij}))_{1\leq j\leq 
    m_i}$ are sampled independently as $X_i(t_{ij})\sim\mathcal{N}(\mu=0,\sigma^2=1)$ when
  $t_{ij}< t^s_i$ (before the change point) and 
  $X_i(t_{ij})\sim\mathcal{N}(\mu=\mu_i,\sigma^2=1)$ when $t_{ij}\geq t^s_i$ (after the
  change point). See Figure \ref{Fig:jump2} for an example;
\item a slope shift: in this final case, the parameter of the shift is a slope
  $\lambda_i$ with $\lambda_i\sim\mathcal{U}([0.02,3])$. Given $\lambda_i$, the
  $(X_i(t_{ij}))_{1\leq j\leq 
    m_i}$ are sampled independently as $X_i(t_{ij})\sim\mathcal{N}(\mu=0,\sigma^2=1)$ when
  $t_{ij}< t^s_i$ (before the change point) and 
  $X_i(t_{ij})\sim\mathcal{N}(\mu=\lambda_i(t_{ij}-t^s_i),\sigma^2=1)$ when $t_{ij}\geq t^s_i$ (after the
  change point). See Figure \ref{Fig:trend2} for an example.
\end{enumerate}
We generate according to this procedure two data sets with 6000 observations
corresponding to 3000 observations with no anomaly, and 1000 observations for
each of the three types of anomalies. The only difference between data set $A$
and data set $B$ is the amplitude of the mean shift which is smaller in $B$,
making the classification harder.

\subsection{Indicators}
\begin{figure}[htbp]
\centerline{\includegraphics[width=0.9\linewidth]{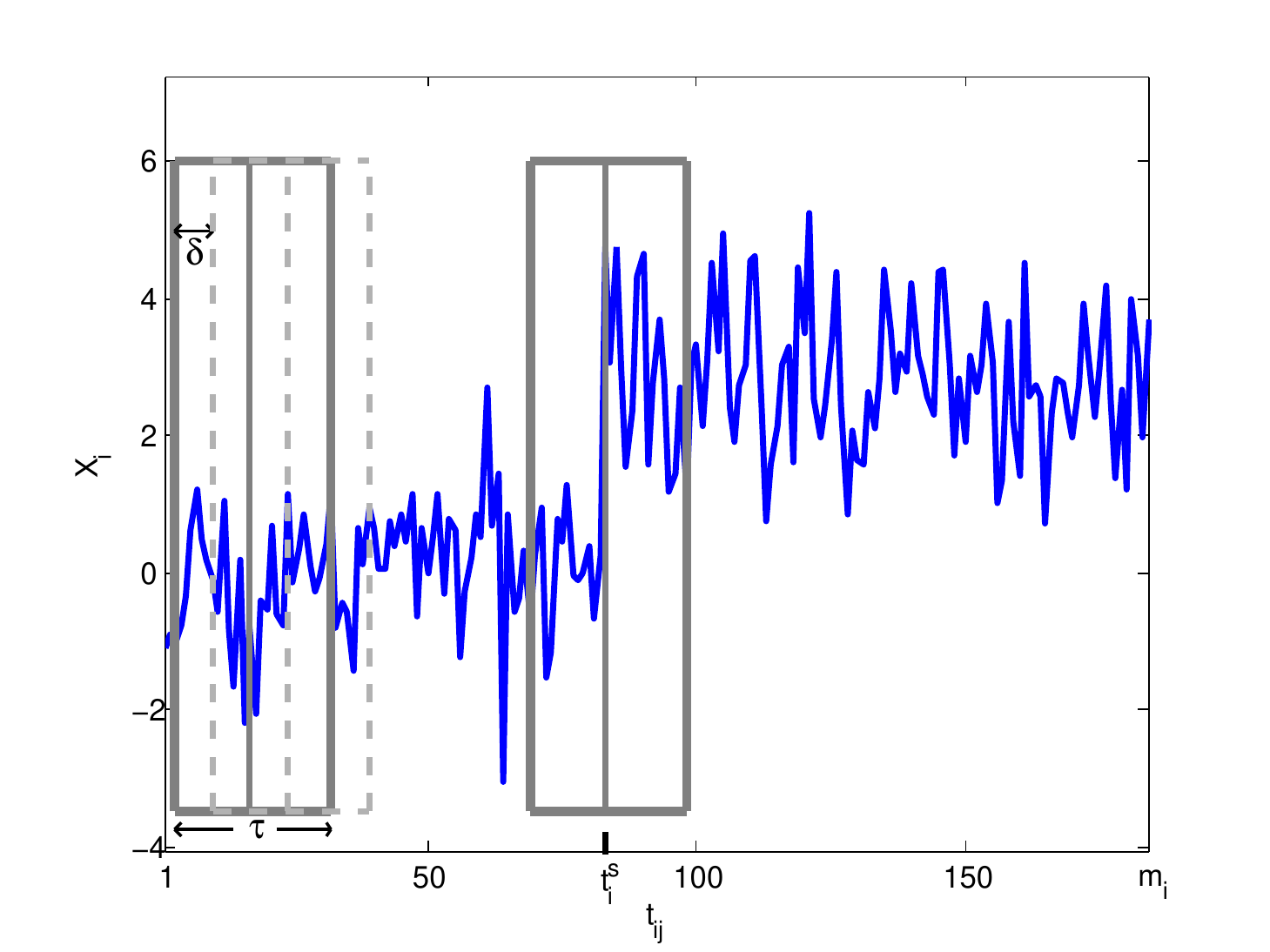}}
\caption{Illustration of the  sliding and jumping windows. $\tau$ is the length of the window. $\delta$ is the jump parameter. $t_i^s$ is the change point.}
\label{fig:sliding}
\end{figure}

As explained in Section \ref{sec:from-anomaly-types}, binary indicators are
constructed from expert knowledge by varying parameters, including scale and
position parameters. In the present context, we use sliding and jumping
windows: for each possible position of the window, a classical statistical
test is conducted to decide whether a shift in the signal occurs at the center
of the window (see Figure \ref{fig:sliding}). Individual tests obtained from different positions are combined
to create binary indicators. 

More precisely, a window of length $\tau$ is a series of $\tau$ consecutive
time points in a signal $(X_i(t_{ij}))_{1\leq j\leq m_i}$ (in other words, this
is a sub-signal). For a fixed $\tau\leq m_i$, there are $m_i-\tau+1$ windows
in $X_i$, from $(X_i(t_{ij}))_{1\leq j\leq \tau}$ to
$(X_i(t_{ij}))_{m_i-\tau+1\leq j\leq m_i}$ (when sorted in order of their first
time points).

Given a window of length $\tau$ (assumed even) starting at position $k$, a two
sample test is conducted on the two subsets of values corresponding to the
first half of the window and to the second half. More precisely, we extract
from the series of values $(X_i(t_{ij}))_{k\leq j\leq k+\tau-1}$ a first
sample $S^1=(X_i(t_{ij}))_{k\leq j\leq k+\tau/2-1}$ and a second sample
$S^2=(X_i(t_{ij}))_{k+\tau/2\leq j\leq k+\tau-1}$. Then a test of inequality
between $S^1$ and $S^2$ is conducted (inequality is defined with respect to some
statistical aspect).  The ``expert'' designed tests are here (notice that
those tests do not include a slope shift test):
\begin{enumerate}
\item the Mann-Whitney-Wilcoxon U test (non parametric test for shift in
  mean);
\item the two sample Kolmogorov-Smirnov test (non parametric test for
  differences in distributions);
\item the F-test for equality of variance (parametric test based on a Gaussian
  hypothesis). 
\end{enumerate}
From this general principle, we derive a large set of indicators by varying
the length of the window, the level of significance of the test and the way
to combine results from all the windows that can be extracted from a signal. 

In practice, for an observation $(X_i(t_{ij}))_{1\leq j\leq m_i}$, we use
three different window lengths, $\tau=30$, $\tau=50$ and $\tau=100$. We use
also three different levels for the tests, namely 0.005, 0.1 and 0.5. For a
fixed window length $\tau$ and a fixed test, an observation
$(X_i(t_{ij}))_{1\leq j\leq m_i}$ is transformed into $m_i-\tau+1$ $p$-values
produced by applying the test to the $m_i-\tau+1$ windows extracted from the
observation. For each significance level, the $p$-values are binarized giving
1 or 0 whether the null hypothesis of identical distribution between $S^1$ and
$S^2$ is rejected or not, leading to $m_i-\tau+1$
binary values.

The next step consists in turning the raw binary values into a set of
indicators producing the same number of indicators for all observations. The
simplest binary indicator equals to 1 if and only if at least one binary value
among the $m_i-\tau+1$ is equal to 1 (that is if at least one window defines
two sub-samples that differ according to the chosen test). Notice that as we
use 3 window lengths, 3 tests, and 3 levels, we obtained this way 27 simple
binary indicators.

Then, more complex binary indicators are generated, as explained in Section
\ref{sec:conf-indic}. In a way, this corresponds to build very simple
binary classifiers on the binary values obtained from the tests. All those
indicators are based on the notion of consecutive windows. In order to vary
the time resolution of this process, we first introduce a jump parameter
$\delta$. Two windows are consecutive according to $\delta$ if the first time
point of the second window is the $(\delta+1)$-th time point of the first
window. For instance, if $\delta=5$, $(X_i(t_{ij}))_{1\leq j\leq \tau}$ and
$(X_i(t_{ij}))_{6\leq j\leq \tau+5}$ are consecutive windows. In practice, we
use three values for $\delta$, namely 1, 5 and 10. 

For each value of $\delta$ and for each series of $m_i-\tau+1$ binary values,
we compute the following derived indicator:
\begin{enumerate}
\item the \emph{global ratio indicator} is equal to 1 if and only if on a fraction
  of at least $\beta$ of all possible windows, the test detects a change. This indicator
  does not use the fact that windows are consecutive, but it is nevertheless
  affected by $\delta$. Indeed, values strictly larger than 1 for $\delta$
  reduce the total number of windows under consideration;
\item the \emph{consecutive ratio indicator} is equal to 1 if and only if there is a
  series of $\beta(m_i-\tau+1)$ \emph{consecutive} windows on which the test detects a change ;
\item the \emph{local ratio indicators} is equal to 1 if and only if there is
  a series of $l$ \emph{consecutive} windows among which the tests detects at least $k$
  times a change.
\end{enumerate}
Those derived indicators are parametric. In the present experiments, we use
three different values for $\beta$ (used by the first two derived indicators),
namely 0.1, 0.3 and 0.5. For the pair $(l,k)$ used by the last derived
indicator, we used three different pairs $(3,2)$, $(5,3)$ and $(5,4)$. 

A simple series of $m_i-\tau+1$ binary values allows us to construct 27
derived indicators (3 values for $\delta$ times $3$ values for the parameters
of each of the 3 types of derived indicators). As we have 27 of such series
(because of the 3 window lengths, 3 tests and 3 levels), we end up with 729
additional indicators for a total of 756 binary indicators. 

In addition, based on expert recommendation, we apply the entire processing 
both to the original signal and to a smoothed signal (using a simple moving
average over 5 measurements). The total number of indicators is therefore
1512. However many of them are identical over the 6000 observations and the
final number of distinct binary indicators is 810. The redundancy is explained
by several aspects. For instance, while the smoothing changes the signal, it has a
limited effect on the test results. Also, when the level of the test is high, the
base binary values tend to be all equal to one. When the ratio $\beta$ is low,
there are not many differences between the derived indicators with respect to
$\delta$, etc. 

It should be noted that the parameters used both for the simulated data and
the indicators have been chosen so as to illustrate the possibilities of the
proposed architecture of the decision process. The values have been considered
reasonable (see Table \ref{tab:param} for a summary of these values) and representative of what would be useful in practice by experts
of our application domain. It is however expected than more statistical tests
and more indicators should be considered in practice to cover the range of the
possible anomalies.

\begin{table}
  \centering
  \begin{tabular}{p{0.44\linewidth}l}
\toprule
Parameters & Values \\
\midrule
\multirow{3}{*}{statistical test} & Mann-Whitney-Wilcoxon U test\\
&Two sample Kolmogorov-Smirnov test\\
&F-test\\
\midrule[0.1pt]
\multirow{3}{*}{length of window ($\tau$)} & 30\\
&50\\
&100 \\
\midrule[0.1pt]
\multirow{3}{*}{levels}  &0.005\\
&0.1\\
& 0.5\\
\midrule[0.1pt]
\multirow{3}{*}{jump ($\delta$)} & 1\\
& 5\\
& 10 \\
\midrule[0.1pt]
\multirow{3}{\linewidth}{fraction for \textit{global ratio indicator} and
    \textit{consecutive ratio indicator} ($\beta$)}  & 0.1\\
&0.3\\
& 0.5 \\
\midrule[0.1pt]
\multirow{3}{\linewidth}{k among l for \textit{local ratio indicators} ($l,k$)}  &
                                                                          (3,2)\\
& (5,3)\\
& (5,4) \\
\midrule[0.1pt]
\multirow{2}{*}{moving average} & 1\\
&5 \\
\bottomrule
  \end{tabular}
\medskip
  \caption{Listing of the values used for the parameters of the indicators.}
  \label{tab:param}
\end{table}

\subsection{Performance analysis}
Each data set is split in a balanced way into a learning set with 1000 signals
and a test set with 5000 signals (the class proportions from the full data set
are kept in the subsets). We report the global classification accuracy (the
classification accuracy is the percentage of correct predictions, regardless
of the class) on the learning set to monitor possible over fitting. The
performances of the methodology are evaluated on 10 balanced subsets of size
500 from the 5000 signals' test set. This allows to evaluate both the average
performances and their variability. For the Random Forest, we also report the
out-of-bag (oob) estimate of the classification accuracy: this quantity is
obtained during the bootstrap procedure used to construct the forest. Indeed
each observation appears as a training observation in only roughly two third
of the trees that constitute the forest. Then the prediction of the remaining
trees for this observation can be aggregated to give a decision. Comparing
this decision to the true value gives the out-of-bag estimate of the
classification error for this observation (see \cite{breiman2001random} for
details). Finally, for the Naive Bayes classifier, we use confusion matrices
and class specific accuracies to gain more insights on the results when
needed.

\subsection{Performances with all indicators}
As indicators are expertly designed and should cover the useful
parameter range of the tests, it is assumed that the best classification
performances should be obtained when using all of them, up to the effects of
the curse of dimensionality. 

\begin{table}
  \centering
  \begin{tabular}{lccc}
\toprule
Data set & Training set acc. & OOB acc. & Test set average acc. \\
\midrule[0.1pt]
$A$ & 0.9770 & 0.9228 &0.9352 (0.0100) \\
$B$ &  0.9709 & 0.9118 & 0.9226 (0.0108) \\
\bottomrule
  \end{tabular}
\medskip
  \caption{Classification accuracy of the Random Forest using the 810 binary
    indicators. For the test set, we report the average classification
    accuracy and its standard deviation between parenthesis.}
  \label{tab:fullRF}
\end{table}

Table \ref{tab:fullRF} reports the global classification accuracy of the
Random Forest, using all the indicators. As expected, Random Forests suffer
neither from the curse of dimensionality nor from strong over fitting (the
test set performances are close to the learning set ones). Table
\ref{tab:fullNBN} reports the same performance indicator for the Naive Bayes
classifier. Those performances are significantly lower than the one obtained
by the Random Forest. As shown by the confusion matrix on Table
\ref{tab:conf:NBNsetAtest}, the classification errors are not concentrated on
one class (even if the errors are not perfectly balanced). This tends to
confirm that the indicators are adequate to the task (this was already obvious
from the Random Forest).

\begin{table}[htbp]
  \centering
  \begin{tabular}{lccc}
\toprule
Data set & Training set accuracy & Test set average accuracy \\
\midrule[0.1pt]
$A$ &0.7856   & 0.7718 (0.0173) \\
$B$ & 0.7545 & 0.7381 (0.0178) \\
\bottomrule
  \end{tabular}
\medskip
  \caption{Classification accuracy of the Naive Bayes classifier using the 810 binary
    indicators. For the test set, we report the average classification
    accuracy and its standard deviation between parenthesis.}
  \label{tab:fullNBN}
\end{table}

\begin{table}
\centering
\begin{tabular}{clccccc}%\begin{tabular}{cl|cccc|c}
\toprule
&&\multicolumn{4}{c}{Predicted class}&\\ %&&\multicolumn{4}{c|}{Predicted class}&\\
&&No anomaly&Variance&Mean&Slope&total\\
\midrule[0.1pt]
\multirow{4}{*}{True class}&No anomaly&1759&667&45&29&2500\\

&Variance&64&712&50&3&829\\

&Mean&7&2&783&37&829\\

&Slope&32&7&195&595&829\\
\bottomrule

\end{tabular}
\caption{Data set $A$: confusion matrix with all indicators for Naive Bayes
  classifier on the full test set.}
\label{tab:conf:NBNsetAtest}
\end{table}

\subsection{Feature selection}
While the Random Forest gives very satisfactory results, it would be
unacceptable for human operators as it operates in a black box way. While the
indicators have simple interpretation, it would be unrealistic to ask to an
operator to review 810 binary values to understand why the classifier favors
one class over the others. In addition, the performances of the Naive Bayes
classifier are significantly lower than those of the Random Forest one. Both
drawbacks favor the use of a feature selection procedure. 

As explained in Section \ref{sec:decision}, the feature selection relies on
the mRMR ranking procedure. A forward approach is used to evaluate how many
indicators are needed to achieve acceptable predictive performances. Notice
that in the forward approach, indicators are added in the order given by mRMR
and then never removed. As mRMR takes into account redundancy between the
indicators, this should not be a major issue. Then for each number of
indicators, a Random Forest and a Naive Bayes classifier are constructed and
evaluated. 

\begin{figure}[htbp]
\centerline{\includegraphics[width=0.9\linewidth]{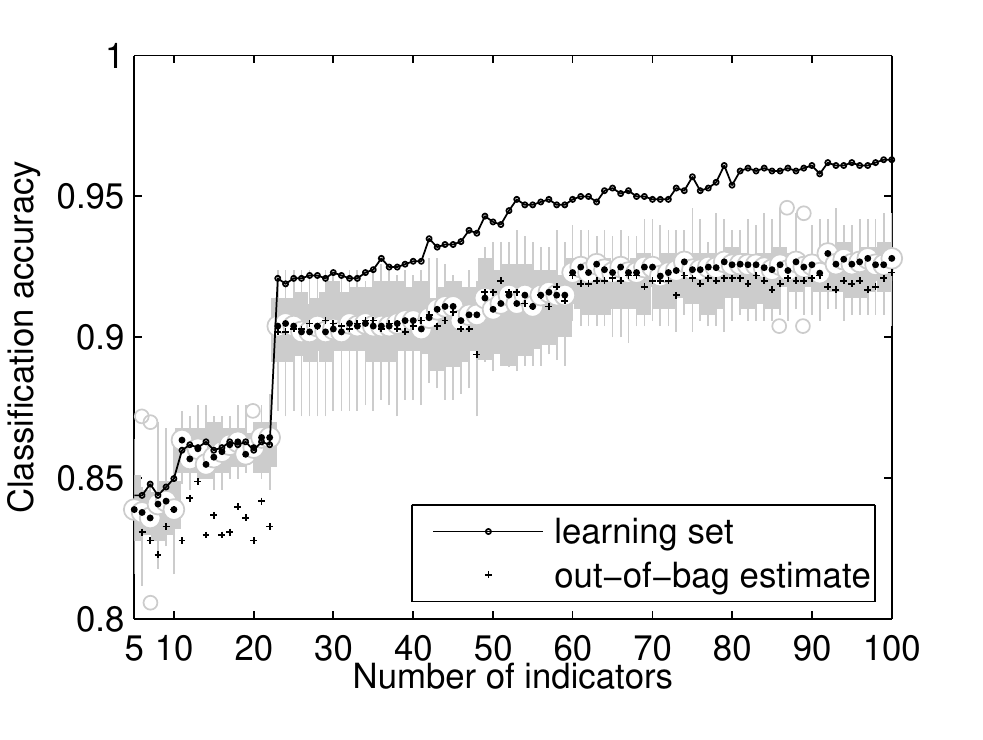}}
\caption{\textbf{Data set $A$ Random Forest}: classification accuracy on learning set (circle) as a function of the
  number of indicators. A boxplot gives the classification accuracies on the
  test subsets, summarized by its median (black dot inside a white
  circle). The estimation of those accuracies by the out-of-bag (oob) bootstrap
  estimate is shown by the crosses.}
\label{fig:rfboxsetA}
\end{figure}

\begin{figure}[htbp]
\centerline{\includegraphics[width=0.9\linewidth]{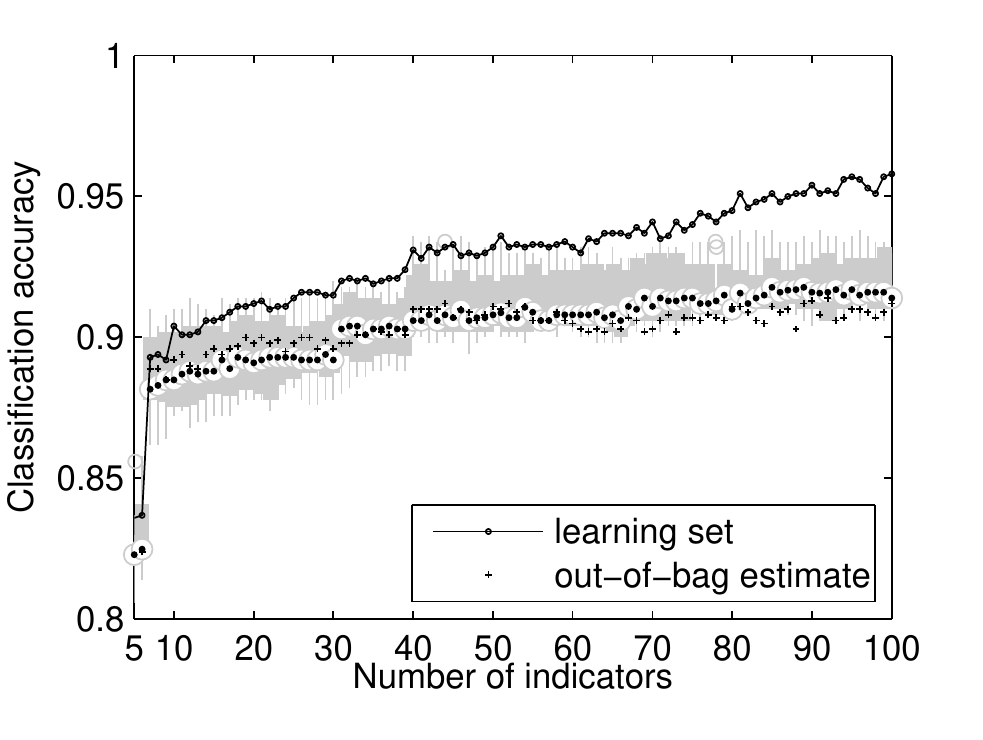}}
\caption{\textbf{Data set $B$ Random Forest}, see Figure \ref{fig:rfboxsetA}
  for details.}
\label{fig:rfboxsetD}
\end{figure}

\begin{figure}
\centering
\includegraphics[width=0.9\linewidth]{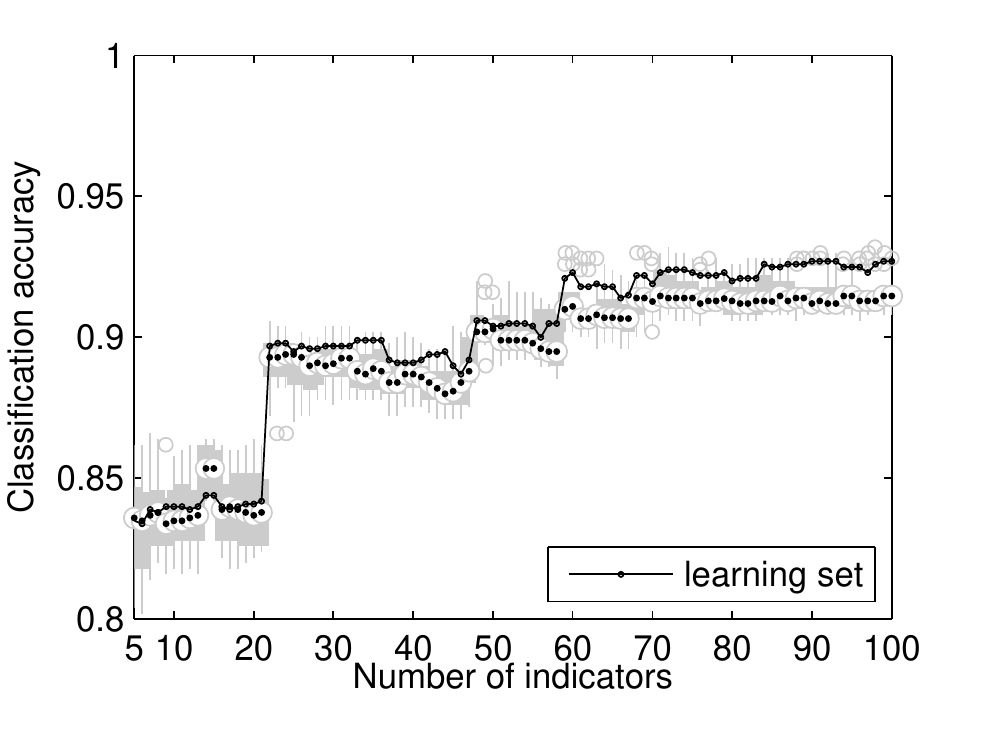}
\caption{\textbf{Data set $A$ Naive Bayes classifier}: classification accuracy on learning set (circle) as a function of the
  number of indicators. A boxplot gives the classification accuracies on the
  test subsets, summarized by its median (black dot inside a white
  circle). }
\label{fig:nbnboxsetA}
\end{figure}

\begin{figure}
\centering
\includegraphics[width=0.9\linewidth]{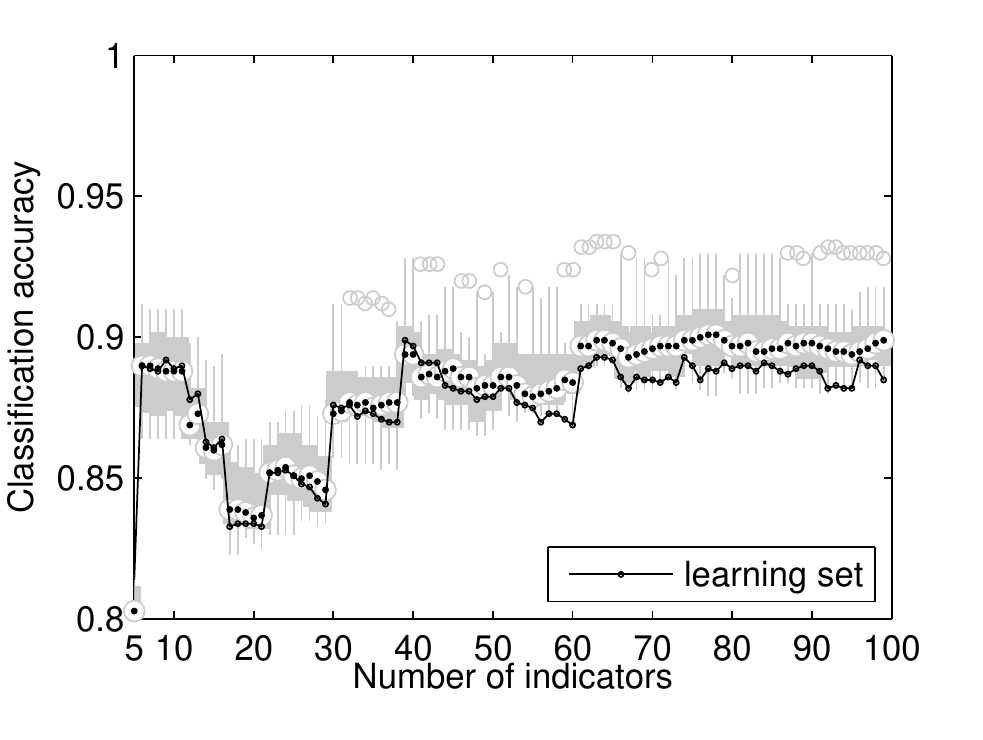}
\caption{\textbf{Data set $B$ Naive Bayes classifier}, see Figure
  \ref{fig:nbnboxsetA} for details.}
\label{fig:nbnboxsetD}
\end{figure}

Figures \ref{fig:rfboxsetA}, \ref{fig:rfboxsetD}, \ref{fig:nbnboxsetA} and
\ref{fig:nbnboxsetD} summarize the results for the 100 first indicators. The
classification accuracy of the Random Forest increases almost monotonously
with the number of indicators, but after roughly 25 to 30 indicators
(depending on the data set), performances on the test set tend to stagnate
(this is also the case of the out-of-bag estimate of the performances, which
shows, as expected, that the number of indicators could be selected using this
measure). In practice, this means that the proposed procedure can be used to
select the relevant indicators implementing this way an automatic tuning
procedure for the parameters of the expertly designed scores. 

Results for the Naive Bayes classifier are slightly more complex in the case
of the second data set, but they confirm that indicator selection is
possible. Moreover, reducing the number of indicators has here a very positive
effect on the classification accuracy of the Naive Bayes classifier which
reaches almost as good performances as the Random Forest. Notice that the
learning set performances of the Naive Bayes classifier are almost identical
to its test set performances (which exhibit almost no variability over the
slices of the full test set). This is natural because the classifier is based
on the estimation of the probability of observing a 1 value
\emph{independently} for each indicator, conditionally on the class. The
learning set contains at least 250 observations for each class, leading to a
very accurate estimation of those probabilities and thus to very stable
decisions. In practice one can therefore select the optimal number of
indicators using the learning set performances, without the need of a
cross-validation procedure (optimality is with respect to the classification
accuracy). 

\begin{figure}
\centering
\includegraphics[width=0.9\linewidth]{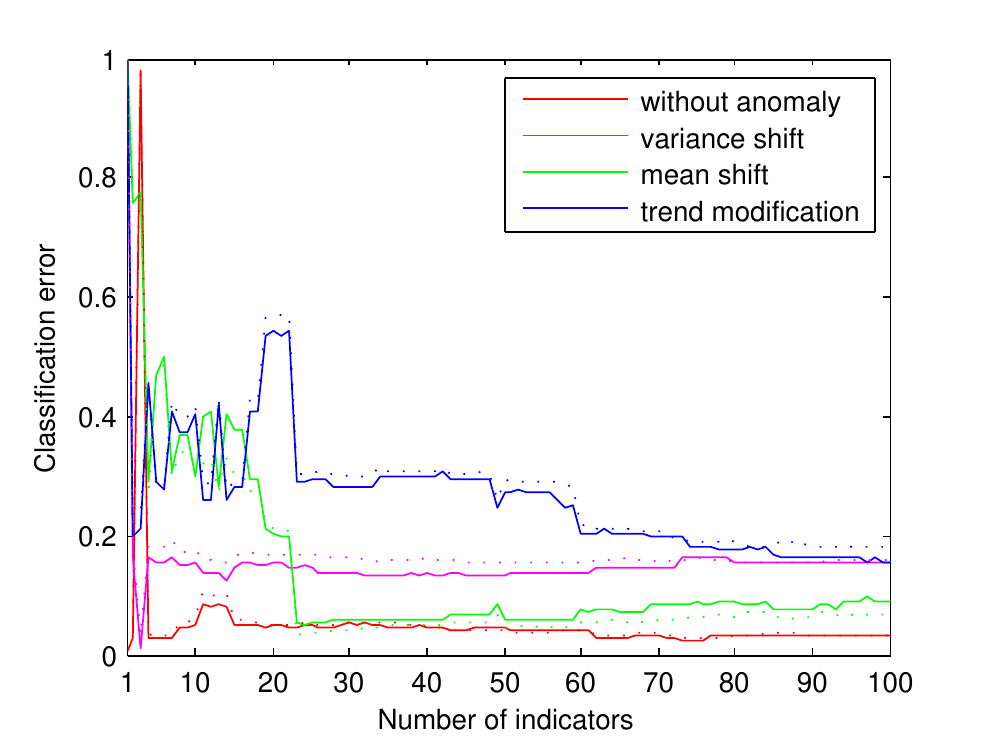}
\caption{\textbf{Data set $A$ Naive Bayes classifier}: classification error
  for each class on the training set (solid lines) and on the test set (dotted
  lines, average accuracies only).}
\label{fig:nbperclassA}
\end{figure}

It should be noted that significant jumps in performances can be observed in
all cases. This might be an indication that the ordering provided by the mRMR
procedure is not optimal. A possible solution to reach better indicator
subsets would be to use a wrapper approach, leveraging the computational
efficiency of both Random Forest and Naive Bayes construction. Meanwhile
Figure \ref{fig:nbperclassA} shows in more detail this phenomenon by
displaying the classification error class by class, as a function of the
number of indicators, in the case of data set $A$. The figure shows the
difficulty of discerning between mean shift and trend shift (for the latter,
no specific test have been included, on purpose). But as the strong decrease in
classification error when the 23-th indicator is added concerns both classes
(mean shift and trend shift), the ordering provided by mRMR could be
questioned. 

\subsection{Indicator selection}
Based on results shown on Figures \ref{fig:nbnboxsetA} and
\ref{fig:nbnboxsetD}, one can select an optimal number of binary indicators,
that is the number of indicators that maximizes the classification accuracy on
the learning set. However, this leads in general to a too large number of
indicators. Thus we restrict the search below a maximal number of indicators
in order to avoid flooding the human operator with to many results. 

For instance Table \ref{tab:NB:selected} gives
the classification accuracy of the Naive Bayes classifier using the optimal
number of binary indicators between 1 and 30. 

\begin{table}[htbp]
\centering
\begin{tabular}{lcccc}
\toprule
Data set & Training set acc. & Test set average acc. & \# of
indicators\\
\midrule[0.1pt]
$A$ &0.8958 & 0.8911 (0.0125) & 23\\
$B$ &0.8828 & 0.8809 (0.0130) & 11\\ 
\bottomrule
\end{tabular}
\caption{Classification accuracy of the Naive Bayesian network using the
optimal number binary indicators between 1 and 30. For the test set, we report the average
classification accuracy and its standard deviation between parenthesis.}
\label{tab:NB:selected}
\end{table}

While the performances are not as good as the ones of the Random Forest, they
are much improved compared to the ones reported in Table \ref{tab:fullNBN}. In
addition, the selected indicators can be shown to the human operator together
with the estimated probabilities of getting a positive result from each
indicator, conditionally on each class, shown on Table
\ref{tab:tenbest:A}. For instance here the first selected indicator,
$confu(2,3)$, is a confirmation indicator for the U test. It is positive when
there are 2 windows out of 3 consecutive ones on which a U test was
positive. The Naive Bayes classifier uses the estimated probabilities to reach
a decision: here the indicator is very unlikely to be positive if there is no
change or if the change is a variance shift. On the contrary, it is very
likely to be positive when there is a mean or a trend shift. While the table
does not ``explain'' the decisions made by the Naive Bayes classifier, it
gives easily interpretable hints to the human operator.  

\begin{table}[htbp]

\centering

\begin{tabular}{ccccc}
\toprule
type of indicator& no change &variance&mean&trend\\
\midrule[0.1pt]

confu(2,3)&0.010333&0.011&0.971&0.939\\

F test&0.020667&0.83&0.742&0.779\\

U test&0.027333&0.03&0.977&0.952\\

ratef(0.1)&0.0016667&0.69&0.518&0.221\\

confu(4,5)&0.034333&0.03&0.986&0.959\\

confu(3,5)&0.0013333&0.001&0.923&0.899\\

U test&0.02&0.022&0.968&0.941\\

F test&0.042&0.853&0.793&0.813\\

rateu(0.1)&0.00033333&0.001&0.906&0.896\\

confu(4,5)&0.019&0.02&0.946&0.927\\

conff(3,5)&0.052333&0.721&0.54&0.121\\

U test&0.037667&0.038&0.983&0.951\\

KS test&0.016&0.294&0.972&0.936\\

confu(3,5)&0.049&0.043&0.988&0.963\\

F test&0.030667&0.841&0.77&0.801\\

U test&0.043&0.043&0.981&0.963\\

lseqf(0.3)&0.0093333&0.749&0.59&0.36\\

rateu(0.1)&0.001&0.002&0.896&0.895\\

lsequ(0.1)&0.062667&0.06&0.992&0.949\\

confu(3,5)&0.025667&0.021&0.963&0.936\\

lseqf(0.3)&0.008&0.732&0.656&0.695\\

KS test&0.016333&0.088&0.955&0.93\\

confu(3,5)&0&0&0.003&0.673\\
\bottomrule

\end{tabular}

\caption{Probability of observing 1 conditionally the class, for each of the  23 best indicators according to mRMR for data set
  $A$. Confu(k,n) corresponds to a positive Mann–Whitney–Wilcoxon U test on
  k windows out of n consecutive ones. Conff(k,n) is the same thing for the
  F-test. Ratef($\beta$) corresponds to a positive F-test on $\beta\times m$
  windows out of $m$. Lseqf($\beta$) corresponds to a positive F-test on
  $\beta\times m$ consecutive windows out of $m$. Lsequ($\beta$) is the same for a U test.
  Detailed parameters of the indicators have been omitted for brevity.}

\label{tab:tenbest:A}
\end{table}

\section{Conclusion and perspectives}
In this paper, we have introduced a diagnostic methodology for engine health
monitoring that leverages expert knowledge and automatic classification. The
main idea is to build from expert knowledge parametric anomaly scores
associated to range of plausible parameters. From those scores, hundreds of
binary indicators are generated in a way that covers the parameter space as
well as introduces simple aggregation based classifiers. This turns the
diagnostic problem into a classification problem with a very high number of
binary features. Using a feature selection technique, one can reduce the
number of useful indicators to a humanly manageable number. This allows a
human operator to understand at least partially how a decision is reached by
an automatic classifier. This is favored by the choice of the indicators which
are based on expert knowledge and on very simple decision rules. A very
interesting byproduct of the methodology is that it can work on very
different original data as long as expert decision can be modelled by a set of
parametric anomaly scores. This was illustrated by working on signals of
different lengths.  

The methodology has been shown sound using simulated data. Using a reference
high performance classifier, Random Forests, the indicator generation
technique covers sufficiently the parameter space to obtain a high
classification rate. Then, the feature selection mechanism (here a simple
forward technique based on mRMR) leads to a reduced number of indicators (23
for one of the data set) with good predictive performances when paired with a
simpler classifier, the Naive Bayes classifier. As shown in the experiments,
the class conditional probabilities of obtaining a positive value for those
indicators provide interesting insights on the way the Naive Bayes classifier
takes a decision. 

In order to justify the costs of collecting a sufficiently large real world
labelled data set in our context (engine health monitoring), additional
experiments are needed. In particular, multivariate data must be studied in
order to simulate the case of a complex system made of numerous
sub-systems. This will naturally lead to more complex anomaly models. We also
observed possible limitations of the feature selection strategy used here as
the performances displayed abrupt changes during the forward procedure. More
computationally demanding solutions, namely wrapper ones, will be studied to
confirm this point. 

It is also important to notice that the classification accuracy is not the
best way of evaluating the performances of a classifier in the engine health
monitoring context. Firstly, engine health monitoring involves intrinsically a strong
class imbalance \cite{japkowicz2002class}. Secondly, engine health monitoring is a
cost sensitive area because of the strong impact on airline profit of an
unscheduled maintenance. It is therefore important to take into account
specific asymmetric misclassification cost to get a proper performance
evaluation.

The Gaussian assumption made on the noise of the simulated data is a strong one. 
In our future work, we plan to use more realistic noise and to test the robustness of the methodology.
We will evaluate if we can compensate this new complexity with the use of more complex indicators.

\bibliography{biblio}
\bibliographystyle{plain}

\end{document}